\documentclass[runningheads]{llncs}
\usepackage{graphicx}
\usepackage[colorlinks=true, allcolors=blue, pagebackref]{hyperref}
\usepackage{cite}

\usepackage[nohyperlinks, printonlyused, nolist]{acronym}
\usepackage{amsmath}

\begin{acronym}[]
\acro{ai}[AI]{Artificial Intelligence}
\acro{lidar}[LiDAR]{Light Detection and Ranging Sensor}
\acro{sota}[SotA]{state-of-the-art}
\acro{icp}[ICP]{Iterative Closest Point}
\acro{ho3d}[HO-3D-r]{HO-3D-r}
\acro{rl}[RL]{Reinforcement Learning}
\acro{il}[IL]{Imitation Learning}
\acro{dnn}[DNN]{Deep Neural Network}
\acrodefplural{DNN}[DNNs]{Deep Neural Networks}
\acro{pdf}[PDF]{probability density function}
\end{acronym}

\begin{document}

\title{
TrackAgent: 6D Object Tracking\\via Reinforcement Learning
\thanks{We gratefully acknowledge the support of the EU-program
EC Horizon 2020 for Research and Innovation under grant
agreement No. 101017089, project TraceBot, the Austrian
Science Fund (FWF), project No. J 4683, and Abyss Solutions Pty Ltd.}
}
\author{
Konstantin Röhrl\inst{1}
\and
Dominik Bauer\inst{1,2}
\and
Timothy Patten\inst{3}
\and
Markus Vincze\inst{1}
}
\authorrunning{Röhrl et al.}
\institute{
TU Wien, Austria \and
Columbia University, United States \and
Abyss Solutions Pty Ltd, Australia\\
}

\maketitle              
\begin{abstract} 
Tracking an object's 6D pose, while either the object itself or the observing camera is moving, is important for many robotics and augmented reality applications. While exploiting temporal priors eases this problem, object-specific knowledge is required to recover when tracking is lost. Under the tight time constraints of the tracking task, RGB(D)-based methods are often conceptionally complex or rely on heuristic motion models.
In comparison, we propose to simplify object tracking to a reinforced point cloud (depth only) alignment task. This allows us to train a streamlined approach from scratch with limited amounts of sparse 3D point clouds, compared to the large datasets of diverse RGBD sequences required in previous works. We incorporate temporal frame-to-frame registration with object-based recovery by frame-to-model refinement using a reinforcement learning (RL) agent that jointly solves for both objectives. We also show that the RL agent's uncertainty and a rendering-based mask propagation are effective reinitialization triggers.
\end{abstract}
%
%
%


\begin{figure}[t]
    \centering
    \includegraphics[width=\linewidth,trim={2cm 4.05cm 1.5cm 7.15cm},clip]{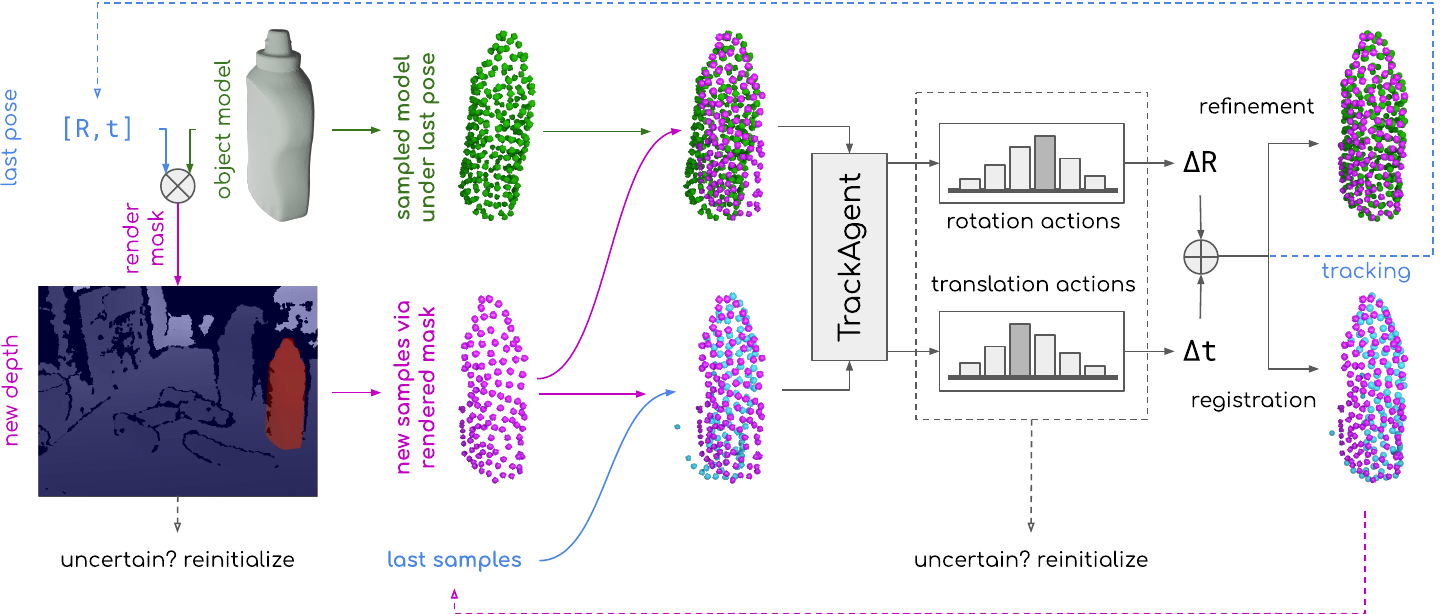}
    \caption{Given a new depth observation, a point cloud segment is extracted by rendering the tracked object under its last known pose. \textit{TrackAgent} then jointly solves two subtasks, namely registration of the observations and pose refinement with respect to the object model. The predicted alignment action iteratively improves the estimated pose, thus tracking an object over multiple frames. If the agent's uncertainty is large, reinitialization is automatically triggered.}
    \label{fig:teaser}
\end{figure}

\section{Introduction}
6D object tracking from sequences of images or video streams has broad applications in computer vision systems. For example, knowing the pose of targets in every frame enables robotic manipulators to grasp objects in dynamic environments~\cite{Marturi2019,Kappler2018,Tuscher2021} likewise, high-frequency poses of obstacles such as pedestrians must be known to plan collision free paths in autonomous driving~\cite{Ess2010_self_autonomous_navigation,Mao2023_survey}.

Multi-frame object tracking exploits prior information in the form of previous estimates. Similar to image or point cloud registration, corresponding information is matched between consecutive frames. However, tracking errors may accumulate and complicate estimation in following frames. 
In comparison, single-frame object pose estimation may recover from such errors by computing the pose from scratch in every frame. Without leveraging temporal information to constrain the potential object poses, however, this becomes a more difficult and computationally expensive problem to solve. This is further exacerbated by an additional refinement step that aligns an object model in the estimated pose to the current frame, e.g., with the \ac{icp} algorithm \cite{icp}.

Both tracking~\cite{se3tracknet,ICG} and pose estimation~\cite{cosypose,liu2022gdrnpp} typically use RGB observations. While textural information allows an object's pose to be accurately determined, RGB features are susceptible to illumination changes and require large volumes of real(istic) training data to learn them.
Depth information, on the other hand, is more robust to changes in texture and illumination. Depth-based features are learned data efficiently \cite{sporeagent} from low-resolution 3D point clouds. Yet, this robustness is paid for by increased ambiguity between similar geometrical shapes, e.g., of multiple box-like objects. Keeping track of the observed segments that correspond to the moving object is therefore complicated.

We observe the complementary nature of tracking via frame-to-frame registration and frame-to-model refinement, as illustrated in Fig.~\ref{fig:teaser}. In our proposed hybrid tracking approach, temporal information is exploited by 1)~registration of corresponding segments between frames and 2)~refinement of the object's model under its previously predicted pose to the current frame's segment. Both objectives are jointly and efficiently solved by a reinforcement learning (RL) agent as a depth-based, point cloud alignment problem. Using a replay buffer to gather the agent's experience during training, the reward may be considered over multiple steps in a single frame or over long time horizons across multiple frames, gaining flexibility over end-to-end supervision. To deal with the ambiguity in the depth modality, we propagate the segmentation information by rendering the mask of the tracked object under its previously predicted pose. Furthermore, our method identifies when the tracked object is lost by considering the confidence of its own estimated pose and mask predictions.

We demonstrate \ac{sota} performance for depth-based object tracking on the YCB-Video dataset \cite{PoseCNN}, closing the gap towards the performance of RGBD-based trackers. We also conduct an ablation study that highlights the efficacy of our hybrid approach over only employing registration and refinement alone as well as alternative subtask fusion methods.


\section{Related Work}
This section discusses related work of registration, that finds a transformation that aligns two point clouds; object pose refinement, which aims to find such a transformation between a (point cloud) observation and a (3D model of an) object; and tracking, which additionally considers object movement.

\subsection{Point Cloud Registration}
The Iterative Closest Point algorithm \cite{icp} is fundamental to solving the registration task. However, \ac{icp} may become stuck in a local optimum and therefore many variants are built around its basic \textit{match-and-update} loop \cite{icpvariants,icpsurvey}.
Seminal work by Aoki et al. \cite{pointnetlk} computes global features per input point cloud using PointNet \cite{pointnet} and proposes a variant of the Lucas-Kanade algorithm to iteratively align the input point clouds. Bauer et al. \cite{bauer2021reagent} train an RL agent to predict these pose updates, treating the global features as its state vector.

Compared to these registration approaches, we propose to additionally consider the refinement towards the tracked object's 3D model to increase robustness to low overlap between the registered point clouds.

\subsection{Object Pose Refinement}
Building upon the RL-based registration in \cite{bauer2021reagent}, SporeAgent \cite{sporeagent} leverages object-related information in the form of symmetries, segmentation labels and physical constraints with respect to multi-object scenes that are jointly refined.
Using RGB images as input instead, DeepIM \cite{deepim} matches an observed and a rendered image of the object under the currently predicted pose to predict refinement transformations. PoseRBPF \cite{poserbpf} pairs a Rao-Blackwellized particle filter with an auto-encoder network. Each particle corresponds to a translation hypothesis. Rotation is determined by comparing each hypotheses image crop to precomputed embeddings for the object under discretized rotations.

Our proposed approach achieves temporal consistency by adding a registration objective to the refinement tasks. Compared to RGB-based refinement approaches, depth-based refinement is found to increase data efficiency \cite{sporeagent} and we further motivate depth-based tracking by robustness to sensor noise since the registration subtask may align point clouds from the same (noisy) distribution.

\subsection{Object Pose Tracking}
Refinement approaches are also used in object pose tracking. By initializing DeepIM \cite{deepim} with the previous frame's pose estimate, tracking performance comparable to dedicated RGBD approaches is achieved. However, since there is no specific means for maintaining object permanence, reinitializations are required when tracking is lost. More adjusted to the tracking task, particles in PoseRBPF \cite{poserbpf} are resampled according to a motion model, which exploits temporal consistency between frames. se(3)-TrackNet~\cite{se3tracknet} proposes a novel feature encoding that improves sim-to-real transfer, thus being able to train on large volumes of synthetic data. The Lie algebra is used to better formulate the loss function for learning pose transformations.
Recently, ICG~\cite{ICG} tracks textureless objects with a probabilistic model. Colour information derives a region-based \ac{pdf} that represents the corresponding lines between a model and its contour in the image plane, while depth information formulates a geometric \ac{pdf}. Optimizing the joint \ac{pdf} finds the pose that best explains the observation (fundamentally similar to ICP~\cite{icp}) to track the object.

Rather than attempting to further extend these already complex tracking pipelines, we follow the simpler approach of using the previous frame's poses for initialization~\cite{deepim} and to propagate segmentation information by rendering. However, we only require depth rendering to crop the observed point cloud and track uncertainty, making textureless 3D models sufficient. By casting tracking as a joint registration and refinement task of 3D point clouds, a slim encoder results in more efficient training than in RGB(D)-based approaches.

\section{TrackAgent: Reinforced Object Pose Tracking}
We consider object pose tracking as a combination of two subtasks. First, the corresponding point cloud observations of an object are aligned between consecutive frames (\textit{registration}). Starting from an initially known pose, the resulting relative transformations yield the object's pose in subsequent time steps. 
Second, the point cloud observation of an object in each individual frame is aligned to its 3D model (\textit{refinement}). Importantly, this subtask enables tracking even when two point cloud observations do not overlap as the object model serves as a common reference frame. In this work we show that the joint consideration of these two subtasks results in better tracking performance over each in isolation.

Fig.~\ref{fig:teaser} gives an overview of our proposed pipeline. In Sec.~\ref{sec:met_agent}, we outline our proposed RL pipeline that solves these two subtasks in a unified manner. In Sec.~\ref{sec:met_mask-reinit} a means to propagate the point cloud observation's segmentation via mask rendering is presented. We moreover discuss heuristics that leverage existing information in our pipeline to automatically trigger a reinitialization when tracking confidence deteriorates. 


\begin{figure}[t]
    \centering
    \includegraphics[width=\linewidth,trim={3cm 7.3cm 1.7cm 4.7cm},clip]{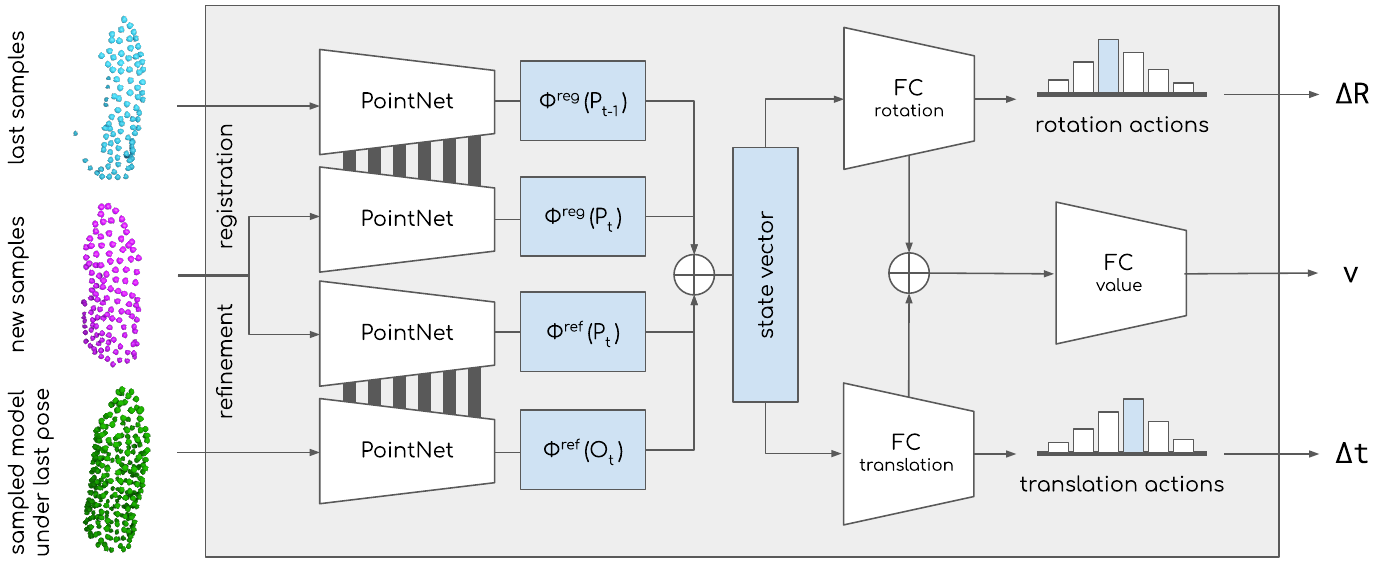}
    \caption{One iteration of \textit{TrackAgent}. Two separate global feature embeddings are learned, one per subtask. These are concatenated to become the state vector of the policy network, which predicts action distributions for rotation and translation. Its goal is to align the current observation more closely with, both, the previous observation and the model. This is repeated for 10 iterations per frame and the final prediction initializes the tracker in the next frame.}
    \label{fig:architecture}
\end{figure}

\subsection{Keeping Track of the Pose: A Reinforced Agent Architecture for Joint Point Cloud Registration and Refinement}\label{sec:met_agent}

\textbf{Defining the Subtasks:} 
Let $D_t$ be the depth image at time~$t$ and $M_t$ the corresponding segmentation mask for the tracked object. We extract a 3D point cloud~$P_t$ from these inputs using the segmented depth and the camera intrinsics. The object is described by its uniformly sampled 3D model~$O$ and pose~$T_t$.

The task of registration is to align $P_t$ to $P_{t-1}$ by a rigid transformation~$T_t^{reg}$. Given an initial pose~$T_0$ and by accumulating all relative registration transformations over~$t \in (0,\mathcal{T}]$, the object's pose $T_\mathcal{T}$ is computed for a given time step~$\mathcal{T}$.
Similarly, we define the task of refinement as finding the alignment by a rigid transformation~$T_t^{ref}$ of $P_t$ to the object's model under some initial pose estimate~$O_t = T_{t-1} \cdot O_{t-1}$.
The combined tracking task that we are aiming to solve is therefore to find a \textit{single} rigid transformation~$T_t^{tra}$ that complies with both subtasks' objectives, i.e., $T_t^{reg} = T_t^{ref} = T_t^{tra}$. This may be viewed as aligning the point clouds from two consecutive frames to one another \textit{as well as} to the common reference frame of the tracked object, i.e., its model.

\textbf{Defining the Agent's Architecture:}
We base our pipeline on related works that treat point cloud alignment as an RL task, previously applied to the registration \cite{bauer2021reagent} and refinement \cite{sporeagent} subtasks. To align two point clouds, these approaches first extract a global feature vector per point cloud with a PointNet-like embedding \cite{pointnet}. Both vectors are concatenated to create a state vector for the alignment policy. Using two separate network branches, discrete distributions are predicted that are interpreted as the policy's rotation and translation actions. The highest-probable actions are selected and the combined transformation constitutes the prediction to more closely align the two point clouds. The agent is trained with a combination of Behavioral Cloning (supervised by expert actions that are derived from the ground-truth transformation) and Proximal Policy Optimization \cite{schulman2017ppo} (guided by a Chamfer distance based reward). We refer readers to the baselines~\cite{bauer2021reagent,sporeagent} for more detailed explanations.

As illustrated in Fig.~\ref{fig:architecture}, for the tracking task, we extend this approach by learning a separate embedding for the registration and refinement objectives, respectively. Two encoders $\phi^{reg}$ and $\phi^{ref}$ are trained, one for embedding point clouds $P_t$ and $P_{t-1}$ and one for embedding $P_t$ and the points sampled on the object model $O_t$ under an initial pose. Specifically, the object model is initialized by the pose of the previous time step $T_{t-1}^{tra}$. The resulting global feature vectors are concatenated and create the new state vector for the joint task. The same policy network as in the baseline approach is used to predict alignment actions based on this state. See Sec.~\ref{sec:ex_hybrid} for a comparison with alternative early (shared embedding) and late fusion (separate policy networks) of subtask information.

\textbf{Defining the Agent's Reward:} The baseline approaches reward the agent for obtaining a closer alignment over multiple steps $i$ in a single frame with respect to the Chamfer distance $d$ between the two point clouds by the reward
\begin{equation}\label{eq:reward}
    r = \begin{cases}
    +0.5,   &\quad d_i < d_{i-1},\\
    -0.1,   &\quad d_i = d_{i-1},\\
    -0.6,   &\quad d_i > d_{i-1},\\
    \end{cases}
\end{equation}
that is, reinforcing a closer alignment, giving a small penalty for pausing steps and a large penalty for diverging steps. The authors of \cite{bauer2021reagent} suggest that using a larger value for diverging than for alignment should discourage the agent from alternating between the two (as a shortcut to increase its reward).

We want to further reinforce the agent to maintain a close alignment over multiple \textit{frames}, not just steps. The intuition is that an alignment that complicates tracking in subsequent frames should be penalized; conversely, an alignment that serves as useful initialization should be reinforced. A straightforward solution to achieve this is to reuse the multi-step reward in Eq.~\eqref{eq:reward} for the final estimates (i.e., final step) in each frame. 
We find that combining both timescales works best, as the dense multi-step reward signal simplifies the training as compared to using a sparse multi-frame reward on its own.

\subsection{Keeping Track of the Observation and the Agent's Uncertainty: Mask Propagation and Automatic Reinitialization Triggers}\label{sec:met_mask-reinit}

\textbf{Propagating and Refining Mask Information:} 
Based on the agent's prediction, we render the corresponding 3D object model under the estimated pose to create the segmentation mask for the following frame. This keeps track of the point cloud segment that corresponds to the object of interest. A segmentation branch, introduced in \cite{sporeagent}, refines the resulting point cloud segment by rejecting outlier points from the max pooling operation of the state embedding. See \cite{sporeagent} for a more detailed discussion of the branch's architecture.

As an additional benefit of the rendering-based mask propagation, we leverage the rendered depth image to compute tracking uncertainty with respect to the observed depth image. A visibility mask is computed that contains all pixels for which the rendered depth is at most $\tau$
behind the observed depth. The number of inlier pixels in the visibility mask is small if there is either a large misalignment (tracking lost) or heavy occlusion (observation lost). In both situations, reinitialization is needed. Relying only on this observed uncertainty, however, results in high sensitivity to heavy occlusion.

\textbf{Combining Uncertainty-based Reinitialization Triggers:}
An alternative source for tracking uncertainty stems from the interpretation of the agent's action distribution. Intuitively, the agent learns to classify the misalignment into discrete steps along each axis and in terms of both translation and rotation. Therefore, in the ideal case where the agent eventually aligns the observations, it should predict a \textit{zero-step} action in the final iteration on each frame. 
We exploit this interpretation by triggering a reinitialization when the predicted misalignment (i.e., alignment stepsize) in the final iteration is consistently large over multiple frames. This is achieved by averaging the predicted stepsizes over multiple frames and selecting the reinitialization threshold such that it is also reached when a large step is predicted for individual frames. Yet, a strategy solely based on the agent's uncertainty may mistrigger reinitializations when either pose or observed points are far from the training distribution, even though the estimated pose might be close to alignment.

We find that, by combining these two strategies, the number of reinitializations is controllably balanced with the achieved tracking performance. In Sec.~\ref{sec:ex_reinit}, we investigate the effects of varying thresholds of the combined approach. 


\section{Experiments}
In this section, we evaluate TrackAgent with established error metrics for object pose estmation and tracking using the publicly available BOP Toolkit~\cite{bop_toolkit} in comparison to other \ac{sota} object tracking methods. Additionally, we perform an ablation study to highlight the relevance of individual components.

\subsection{Procedure}
\textbf{Data:}
The YCB-Video~(YCB-V) dataset~\cite{PoseCNN} consists of 92 video sequences of 21 YCB objects~\cite{Calli2015_YCB}, where a camera moves around a static scene of up to six objects. The official dataset splits this into 80 sequences for training and 12 for testing. In this work, we only use every 7$^{th}$ frame in the 80 sequences for training and evaluate on all of the keyframes in the test set. No synthetic data is used.

\textbf{Metrics:} 
We consider the Average Distance of Model Points (ADD) and ADD with Indistinguishable Views (ADI)~\cite{linemod} as error metrics. ADD measures the mean distance
between corresponding points of the model in the estimated and ground-truth poses, while ADI measures the mean distance between the nearest points. ADI is thus better suited for (geometrically) symmetrical objects. 

\textbf{Implementation Details:}
For alignment actions, the stepsizes are set to $[0.00066, 0.002, 0.006, 0.018, 0.054]$, both in the positive and negative directions and with an additional ``stop''-action (i.e., stepsize $0$). These are set to radians for rotation and to units in the normalized space for translation.

During training, the agent is initialized with the ground-truth pose and learns from trajectories of ten consecutive frames with ten alignment iterations each. We use two separate replay buffers each with a length of 128. The first contains the alignment trajectories in a single frame, while the second replay buffer is used to optimize trajectories over consecutive frames. To simulate inaccurate segmentation, we follow the augmentation strategy introduced in~\cite{sporeagent}, where point clouds are sampled with $80\%$ foreground and $20\%$ background. In each epoch, a new set of trajectories is generated from the training split such that the total number of frames is equal to $1/5^{th}$ of all frames in our already subsampled dataset. This strategy both significantly reduces training time and increases generalization. The model is trained for 60 epochs with the same hyperparameters as in~\cite{sporeagent}.


\begin{table}[t]
\caption{Result on YCB-V~\cite{PoseCNN}. Reported recalls are the AUC (mean recall for equally-spaced thresholds in~$(0,10cm]$) in percent. Results for the RGBD (\ddag) and two of the depth-only baselines (\dag) are taken from~\cite{ICG}. All methods are (re)initialized with ground-truth poses. $\S$ indicates reinitialization every 30 frames. Best overall AUC per metric is highlighted in \textbf{bold}, best depth-based AUC per metric in \textit{italics}.
}
\resizebox{\linewidth}{!}{%
\tiny
\setlength{\tabcolsep}{2pt}
\begin{tabular}{l|ll|ll|ll|ll|ll|ll|ll|ll|ll}
\textbf{AUC [\%] $\uparrow$} &
  \multicolumn{2}{l|}{se(3)$^\ddag$ \cite{se3tracknet}} &
  \multicolumn{2}{l|}{ICG$^\ddag$ \cite{ICG}} &
  \multicolumn{2}{l|}{POT$^\dag$ \cite{wuethrich}} &
  \multicolumn{2}{l|}{RGF$^\dag$ \cite{RGF}} &
  \multicolumn{2}{l|}{ICP$^\S$ \cite{icp,open3d}} &
  \multicolumn{2}{l|}{Reg$^\S$ \cite{bauer2021reagent}} &
  \multicolumn{2}{l|}{Ref$^\S$ \cite{sporeagent}} &
  \multicolumn{2}{l|}{\textbf{ours$^\S$}}  &
  \multicolumn{2}{l}{\textbf{ours}}\\
Metric & ADD  & ADI & ADD  & ADI & ADD   & ADI &  ADD   & ADI & ADD   & ADI & ADD   & ADI  & ADD   & ADI  & ADD   & ADI  & ADD   & ADI \\ \hline
master chef can         & \textbf{93.9} & \textbf{96.3} & 66.4 & 89.7 & 55.6 & 90.7 & 46.2 & 90.2 & 78.7 & \textit{94.1} & \textit{84.2} & 92.2 & 80.2 & 91.6 & 76.8 & 92.0 & 75.0 & 91.1 \\
cracker box         & \textbf{96.5} & \textbf{97.2} & 82.4 & 92.1 & \textit{96.4} & \textit{\textbf{97.2}} & 57.0 & 72.3 & 78.3 & 88.4 & 84.9 & 91.9 & 88.6 & 92.8 & 93.3 & 95.7 & 92.3 & 95.1 \\
sugar box               & \textbf{97.6} & 98.1 & 96.1 & \textbf{98.4} & \textit{97.1} & \textit{97.9} & 50.4 & 72.7 & 81.0 & 92.2 & 89.8 & 94.9 & 95.7 & 97.4 & 95.6 & 97.0 & 95.6 & 97.0 \\
tomato soup can         & \textbf{95.0} & 97.2 & 73.2 & \textbf{97.3} & 64.7 & 89.5 & 72.4 & 91.6 & 75.3 & 91.4 & \textit{92.8} & \textit{95.8} & 90.8 & 94.4 & 91.0 & 95.7 & 87.8 & 92.6 \\
mustard bottle          & 95.8 & 97.4 & 96.2 & \textbf{98.4} & \textit{\textbf{97.1}} & 98.0 & 87.7 & \textit{98.2} & 88.0 & 94.4 & 94.3 & 96.4 & 96.6 & 97.7 & 96.0 & 97.5 & 96.1 & 97.6 \\
tuna fish can           & 86.5 & 91.1 & 73.2 & \textbf{95.8} & 69.1 & \textit{93.3} & 28.7 & 52.9 & 76.4 & 88.1 & \textit{\textbf{90.4}} & \textit{93.3} & 60.0 & 81.9 & 64.5 & 82.6 & 73.6 & 91.7 \\
pudding box             & \textbf{97.9} & \textbf{98.4} & 73.8 & 88.9 & \textit{96.8} & \textit{97.9} & 12.7 & 18.0 & 81.5 & 91.4 & 93.6 & 96.1 & 89.2 & 94.0 & 94.4 & 96.5 & 94.0 & 96.4 \\
gelatin box             & \textbf{97.8} & 98.4 & 97.2 & \textbf{98.8} & \textit{97.5} & \textit{98.4} & 49.1 & 70.7 & 81.3 & 92.0 & 95.7 & 97.1 & 87.6 & 92.9 & 96.9 & 97.7 & 96.9 & 97.7 \\
potted meat can         & 77.8 & 84.2 & \textbf{93.3} & \textbf{97.3} & \textit{83.7} & 86.7 & 44.1 & 45.6 & 81.6 & \textit{89.9} & 75.5 & 80.7 & 77.5 & 80.1 & 78.9 & 82.5 & 83.4 & 88.8 \\
banana                  & 94.9 & 97.2 & \textbf{95.6} & \textbf{98.4} & 86.3 & 96.1 & 93.3 & \textit{97.7} & 71.9 & 87.2 & 88.8 & 92.8 & \textit{94.8} & 97.0 & 93.6 & 96.6 & 93.8 & 96.7 \\
pitcher base            & 96.8 & 97.5 & 97.0 & \textbf{98.8} & 97.3 & 97.7 & \textit{\textbf{97.9}} & \textit{98.2} & 90.3 & 96.1 & 92.9 & 95.9 & 94.5 & 96.9 & 95.3 & 97.1 & 95.4 & 97.2 \\
bleach cleanser         & \textbf{95.9} & 97.2 & 92.6 & \textbf{97.5} & 95.2 & 97.2 & \textit{\textbf{95.9}} & \textit{97.3} & 71.7 & 89.2 & 82.0 & 91.9 & 93.2 & 96.4 & 91.4 & 95.5 & 91.3 & 95.5 \\
bowl                    & \textbf{80.9} & 94.5 & 74.4 & \textbf{98.4} & 30.4 & \textit{97.2} & 24.2 & 82.4 & \textit{78.1} & 93.1 & 72.9 & 85.7 & 64.1 & 87.3 & 48.7 & 78.6 & 63.1 & 88.1 \\
mug                     & 91.5 & 96.9 & \textbf{95.6} & \textbf{98.5} & 83.2 & 93.3 & 60.0 & 71.2 & 81.8 & 93.3 & 84.0 & 89.6 & 93.9 & 96.6 & \textit{94.4} & \textit{96.9} & 94.3 & \textit{96.9} \\
power drill             & 96.4 & 97.4 & 96.7 & \textbf{98.5} & 97.1 & 97.8 & \textit{\textbf{97.9}} & \textit{98.3} & 78.1 & 90.3 & 90.6 & 94.1 & 94.1 & 96.2 & 92.4 & 95.0 & 94.3 & 96.5 \\
wood block              & 95.2 & 96.7 & 93.5 & \textbf{97.2} & \textit{\textbf{95.5}} & \textit{96.9} & 45.7 & 62.5 & 71.9 & 90.5 & 83.6 & 91.6 & 89.4 & 95.6 & 91.3 & 95.6 & 91.4 & 95.6 \\
scissors                & \textbf{95.7} & \textbf{97.5} & 93.5 & 97.3 & 35.6 & \hspace{1.5ex}4.2 & 16.2 & 20.9 & 38.6 & 39.9 & 73.5 & 57.3 & 69.1 & 70.2 & 79.5 & 72.3 & \textit{80.1} & \textit{90.0} \\
large marker            & \textbf{92.2} & 96.0 & 88.5 & \textbf{97.8} & 35.6 & 53.0 & 12.2 & 18.9 & 51.1 & 70.5 & 50.5 & 66.4 & 45.6 & 64.6 & 54.9 & 68.4 & \textit{67.0} & \textit{94.2} \\
large clamp             & \textbf{94.7} & \textbf{96.9} & 91.8 & \textbf{96.9} & 61.2 & 72.3 & 62.8 & 80.1 & 63.6 & 77.6 & 81.0 & 87.5 & \textit{82.1} & 88.6 & \textit{82.1} & 89.0 & 81.9 & \textit{89.7} \\
extra large clamp       & 91.7 & 95.8 & 85.9 & 94.3 & \textit{\textbf{93.7}} & \textit{\textbf{96.6}} & 67.5 & 69.7 & 60.5 & 80.2 & 80.7 & 88.5 & 78.7 & 88.2 & 89.1 & 94.9 & 90.8 & 95.7 \\
foam brick              & 93.7 & 96.7 & 96.2 & \textbf{98.5} & \textit{\textbf{96.8}} & \textit{98.1} & 70.0 & 86.5 & 69.6 & 83.8 & 94.7 & 96.4 & 85.2 & 90.9 & 95.3 & 96.9 & 95.1 & 96.9 \\ \hline
\textbf{All Frames}     & \textbf{93.0} & 95.7 & 86.4 & \textbf{96.5} & 78.0 & 90.2 & 59.2 & 74.3 & 74.9 & 88.6 & 84.6 & 90.6 & 83.6 & 90.7 & 84.9 & 91.6 & \textit{86.8} & \textit{93.6}  \\
\end{tabular}}
\label{tab:ycbv_results}
\end{table}

\subsection{Tracking Performance on YCB-V}

\begin{figure}[t]
    \centering
    \includegraphics[width=0.85\linewidth,trim={2.7cm 5.5cm 1.1cm 3.5cm},clip]{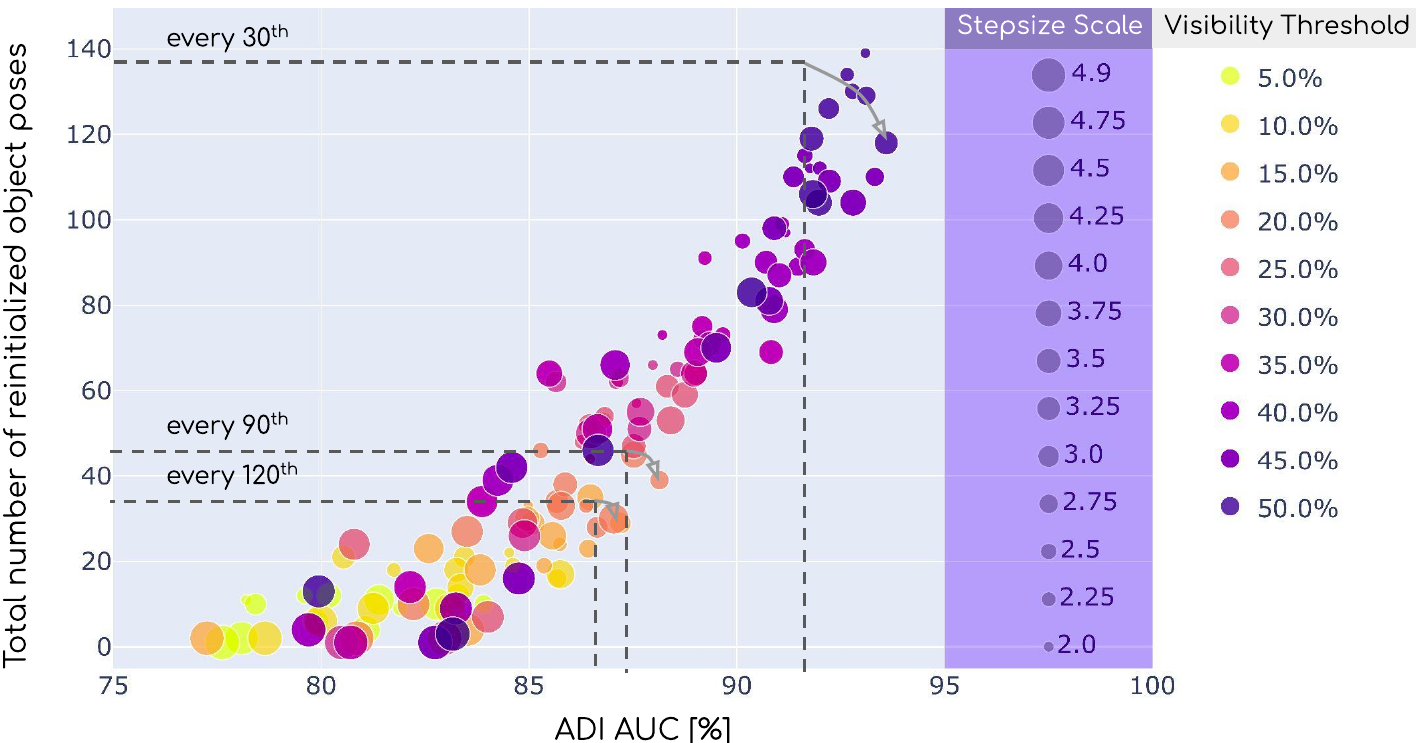}
    \caption{Reinitialization experiments. Methods are grouped by the visibility threshold with the stepsize threshold indicated by the marker size. Reinitializations are counted and carried out per object. High visibility thresholds correspond to a significantly higher number of reinitializations, while a high threshold for the buffered stepsize reduces the number of reinitializations close to zero. For comparison, static reinitialization every $30/90/120^{th}$ frame are shown (dashed). For each value we automatically find thresholds that reduce reinitializations while improving recall (gray arrows). Best viewed digitally.
   }
    \label{fig:reinitialization_action_mask}
\end{figure}

We report the performance of our proposed hybrid (registration and refinement) approach as well as the two subtasks individually on the YCB-V dataset in Table~\ref{tab:ycbv_results}. As can be seen, the hybrid approach achieves higher ADD/ADI recall than agents that consider only one subtask. In comparison to the \ac{icp} and two other depth-based baselines \cite{wuethrich,RGF}, all reinforced implementations (registration \cite{bauer2021reagent}, refinement \cite{sporeagent}, ours) achieve better tracking performance. Moreover, our proposed reinitialization strategy further improves the AUC recall of our approach (right most column), while reducing the number of reinitializations as compared to a fixed time interval (see Fig.~\ref{fig:reinitialization_action_mask}).
The results for \ac{sota} RGBD methods taken from~\cite{ICG} demonstrate that our approach is comparable despite only using the depth modality. The results are particularly similar to ICG~\cite{ICG} and we achieve similar performance compared to both se(3)-TrackNet~\cite{se3tracknet} and ICG~\cite{ICG} for some objects. It should be noted that se(3)-TrackNet exploits a large volume of synthetic data in training where as ours and ICG do not.


\begin{table}[t]
\caption{Ablation study on YCB-V~\cite{PoseCNN} with different state vector fusion-variants.}
\centering
\footnotesize
\begin{tabular}{l|ll|ll|ll}
\begin{tabular}[c]{@{}l@{}}
Reinitialize every n$^{th}$ frame \end{tabular} &
  \multicolumn{2}{c|}{30} &
  \multicolumn{2}{c|}{90} &
  \multicolumn{2}{c}{120} \\
\textbf{AUC [\%]} $\uparrow$ & ADD  & ADI & ADD  & ADI & ADD   & ADI \\ \hline
Registration  & 81.3 & 88.3 & 74.3 & 81.9 & 73.5 & 80.3 \\ 
Refinement  & 83.6 & 90.7 & 78.0& 85.0 & 77.3& 85.1 \\ 
Tracking (early fusion) & 74.0 & 86.0 & 66.5& 77.1 & 63.0 & 73.4 \\
Tracking (late fusion) & 83.7 & 90.2 & 77.2 & 84.3 & 76.4 & 83.3 \\ 
\textbf{TrackAgent} & \textbf{84.9} & \textbf{91.6} & \textbf{80.0}& \textbf{87.3} & \textbf{79.3} & \textbf{86.6} \\ 
\hline
\end{tabular}
\label{tab:hybrid_fusion_ablation}
\end{table}
 
\subsection{Ablation Study}\label{sec:ex_hybrid}
\textbf{Subtask Fusion Variants:}
We compare variants of fusing the subtask information at different points in the agent's architecture. To identify the best combination of registration and refinement information, we explore three milestones in the baseline architecture 1) input point clouds, 2) after state embedding, and 3) after the policy embeddings,
resulting in three proposed architectures.
\begin{itemize}
\item \emph{Early Fusion:}
Merging the input point clouds, the architecture is unchanged and only a larger state dimension is used.
\item \emph{TrackAgent:} See Sec.~\ref{sec:met_agent} and Fig.~\ref{fig:architecture}.
\item \emph{Late Fusion:}
Cloning the state embedding and duplicating the policy network, registration and refinement actions are separately predicted and merged.
\end{itemize}

Table~\ref{tab:hybrid_fusion_ablation} presents the results for the different fusion variants. Even though early fusion is the most lightweight, our experiments show, unsurprisingly, that it is outperformed by more complex strategies. Late fusion achieves similar results to the refinement agent, not being able to leverage the additional temporal information of the registration task. Finally, our TrackAgent is able to outperform agents that consider either subtask alone and both alternative fusion variants. Depending on the number of reinitializations allowed, our approach is able to increase the recall by up to $2.0\%$ ADD AUC and $2.3\%$ ADI AUC compared to the second best variant. 


\textbf{Reinitialization Heuristics:}\label{sec:ex_reinit}
In real-world applications, without annotated data, a reinitialization strategy must not rely on any source of ground-truth, but still trigger when tracking of the object is lost.
A naive way is to reinitialize after a certain time has elapsed, e.g., triggering reinitialization every 30 frames. In comparison, the approach we propose in Sec.~\ref{sec:met_mask-reinit} leverages two measures of uncertainty that are readily available in our tracking pipeline, namely visibility of the object under the predicted pose and the agent's action distribution. Fig.~\ref{fig:reinitialization_action_mask} visualizes an extensive study on how these two correlate and how they affect the achieved tracking recall. Importantly, we see that, as compared to each of the fixed reinitialization time steps, our automatic approach achieves a higher recall using fewer reinitializations. The thresholds for each of the two measures of uncertainty allow the tradeoff between tracking performance and the number of required calls to a reinitialization method to be controlled.\\


\begin{figure}
    \centering
    \includegraphics[width=0.85\linewidth,trim={0cm 2.5cm 0cm 2.8cm},clip]{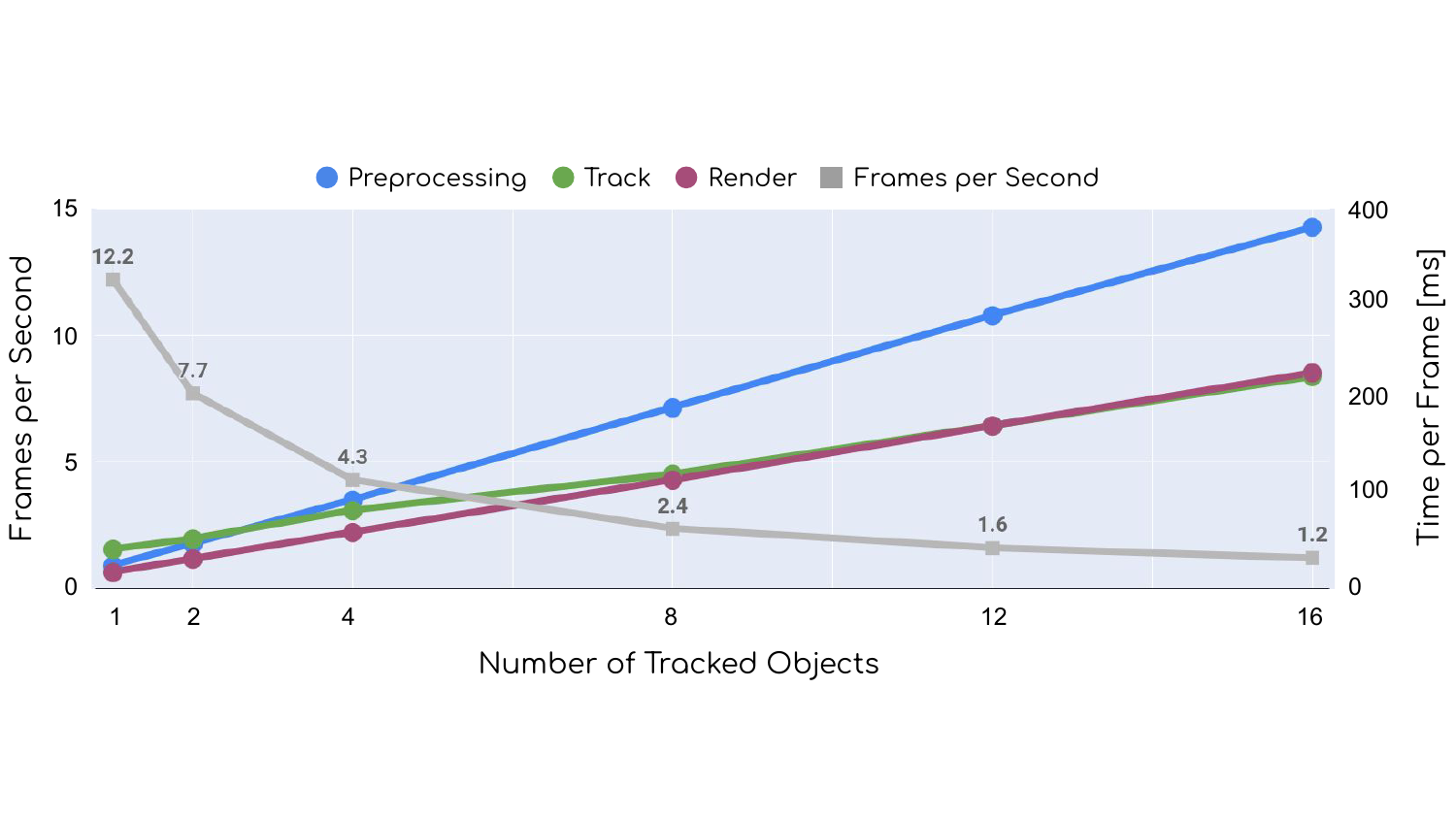}
    \caption{Runtime of TrackAgent for a varying number of objects. Best viewed digitally.
   }
    \label{fig:runtime}
\end{figure}

\textbf{Runtime:}
The runtimes shown in Fig.~\ref{fig:runtime} are measured on a system equipped with an Intel i7-8700K and a NVIDIA Geforce RTX 2080 Ti. As shown, for a single object, the most time is spent on the forward pass of the network (``track''). 
However, parallel processing for multiple objects in a batch scales such that this becomes the fastest part at about 12 objects per frame. Conversely, preprocessing scales linearly and becomes the slowest part at about 4 objects. Optimizing this step, for example the extraction of the point cloud and its transfer to the GPU, has the potential to significantly decrease the runtime. Similarly, for rendering, the bottleneck is reading the rendered image from GPU to CPU memory for preprocessing. While minimal runtime is not the focus for this prototype, we see ample potential for our method in applications with tighter time constraints.


\section{Discussion}
Our experiments demonstrate that our approach using only depth data closes the gap with \ac{sota} RGBD object trackers. Notably, se(3)-TrackNet~\cite{se3tracknet} also leverages a large volume of synthetic data, while we only use a fraction of the real data in the training split.
Our ablation study highlights that jointly solving both subtasks of registration and refinement significantly outperforms each strategy on its own.
By controlling thresholds on two uncertainty metrics, our proposed automatic reinitialization strategy balances the number of reinitializations with tracking accuracy.

While we exploit temporal information through mask propagation and the registration subtask, we expect the inclusion of a motion model to reduce the number of cases where tracking cannot be recovered. On one hand, this could prevent reinitializations that are far outside the training distribution, while still recovering by solving the refinement subtask. Similarly, we observe failure cases where a tracking error translates into decimated segmentation in the following steps, leading to an eventual loss of tracking. Again, accounting for (continuous) motion by enlarging the rendered mask along the movement direction might avoid losing the tracked object from the segmented point cloud. Finally, this would provide a means to ensure object permanence even under heavy occlusion or when the tracked object is outside the camera frame.


\section{Conclusion}
We presented TrackAgent: a reinforcement learning-based object tracking approach that combines registration with refinement. We show experimentally that our joint approach outperforms solving each subtask in isolation. We also demonstrate that the RL agent's uncertainty in its action predictions and mask propagation is an effective trigger for reinitialization. Finally, experiments on the YCB-V dataset show that TrackAgent is able to outperform all compared depth-only baselines and closes the gap to state-of-the-art RGBD trackers.

For future work, inspired by BundleSDF~\cite{bundlesdf}, a promising direction is to consider the subtask of object reconstruction during tracking.
Another avenue for research is the consideration of multiple objects or hand-object interactions in dynamic scenes such as the DexYCB dataset \cite{dexycb}, enabling the benefits of our RL-based tracker to be used in complex environments and handover scenarios.

%
%
\bibliographystyle{splncs04}
\bibliography{references}
\end{document}